# Achieving Explainability for Plant Disease Classification with Disentangled Variational Autoencoders

HABARAGAMUWA Harshana[*], OISHI Yu, TANAKA Kenichi

National Agriculture and Food Research Organization (NARO),

*Corresponding author at: National Agriculture and Food Research Organization (Japan),    E-mail address: h_harshana@affrc.go.jp

## Abstract

Agricultural image recognition tasks are becoming increasingly dependent on systems based on deep learning (DL); however, despite the excellent performance of DL, it is difficult to comprehend the type of logic or features of the input image it uses during decision making. Knowing the logic or features is highly crucial for result verification, algorithm improvement, training data improvement, and knowledge extraction. However, the explanations from the current heatmap-based algorithms are insufficient for the abovementioned requirements. To address this, this paper details the development of a classification and explanation method based on a variational autoencoder (VAE) architecture, which can visualize the variations of the most important features by visualizing the generated images that correspond to the variations of those features. Using the PlantVillage dataset, an acceptable level of explainability was achieved without sacrificing the classification accuracy. The proposed method can also be extended to other crops as well as other image classification tasks. Further, application systems using this method for disease identification tasks, such as the identification of potato blackleg disease, potato virus Y, and other image classification tasks, are currently being developed.

***Keywords*** — *Explainable deep learning; Variational autoencoder; Feature visualization; Disentangled*

*representation; Crop disease classification*

## 1. Introduction

Deep learning (DL) based systems, especially those that use deep convolutional neural networks (DCNNs), are widely used in agricultural image recognition (Habaragamuwa et al., 2018; Kamilaris & Prenafeta-Boldú, 2018). However, once an image has been recognized, the rationale behind recognitions (decisions) performed by the DCNN must be explained, which continues to be an active field of research (Barredo Arrieta et al., 2020).

There are several reasons why the decisions made by a DCNN must be understood. One reason is that a DCNN may unintentionally learn false features (an artifact) during discrimination (Leek et al., 2010; Ribeiro et al., 2016), cognitive biases (annotations made with unreasonable assumptions) (Bolukbasi et al., 2016)-based features, or non-robust features such as textures (Geirhos et al., 2018). Further, by interpreting the features and the relationships between them, key information can be obtained. To address the abovementioned requirements, the type of features and how their variations were used in the DCNN decision making as well as their relative importance must be understood. Hereafter, we refer to these features as important features.

The most popular DCNN algorithms for the visualization of important features describe the area of the image that is the most crucial to the decision making process (such as a heat map) (Montavon et al., 2017; Selvaraju et al., 2020; Toda & Okura, 2019). However, these methods cannot help us identify the exact important features that were used in the classification (Hase et al., 2019) (e.g., color or shape). For example, if yellowing and wrinkling are two features at the same location on a leaf, a heat map cannot be used to identify which features were used in the classification. In addition, although a dataset contains variations of these features (e.g., different degrees of yellowing or wrinkling), heatmaps cannot express these variations. Moreover, heatmaps also cannot express whole-image features, such as image resolution. Therefore, a heat map cannot provide a complete explanation of the features used in the classification as well as their variations.

Our proposed method focuses on developing a feature visualization method based on the variation of features. It shows what the classification system considers as features and the relative importance of those features while maintaining an acceptable classification accuracy.

The rest of this paper is structured as follows: First, the importance of the proposed system and the background is discussed. The second section introduces the proposed method, the technical details, and the dataset details. The third section provides the experimental results and discussions, and the fourth section presents the conclusions of this study.

## 1.2. Background

Explainability is a vital aspect of agricultural image classification; as such, several researchers who have developed agricultural image classification algorithms have attempted to incorporate explainability into their approaches. One of the pioneering approaches to achieve explainability involves the use of activation maps; specifically, these approaches use thresholding to visualize the activation maps (Ghosal et al., 2018; Jiang et al., 2019; Mohanty et al., 2016). However, only visualizing the first activation layer is not sufficient for a classification that involves the top layers of the neural network. Therefore, researchers are now using methods such as saliency and guided backpropagation to achieve explainability (Brahimi et al., 2018). Another approach involves the use of occlusion maps, in which a part of the image is occluded and the changes in the activation maps are observed (Brahimi et al., 2017). To use this method, the user must guess the exact size and shape of the occlusion. In recent years, the Grad-CAM (Selvaraju et al., 2020) algorithm has become popular among researchers in the agricultural field (Desai et al., 2019; Hansen et al., 2018) with some researchers also developing their own approaches to explain the classification process involving a u-net architecture (Brahimi et al., 2019). A good review of such approaches can be found in (Toda & Okura, 2019), with most of these methods focus on visualizing the important areas. However, these visualizations do not provide an idea of how a feature is represented in a neural network.

Numerous attempts based on a variational autoencoder (VAE) approach have been made to find interpretable representations from data, such as infoGan (X. Chen et al., 2016), Beta-VAE (Higgins et al., 2016), and FactorVAE (Kim & Mnih, 2018). Moreover, in recent research (Z. Chen et al., 2020), concept whitening, utilizing a concept whitening layer and a labeled concept dataset, has gained popularity. However, in the current research, the independent features (concepts) and their variations in the image dataset that can be used for the classification of the dataset were identified.

## 2. Proposed Method

To develop the explainable algorithm, a specific architecture was used, where the basic idea was to train a VAE that will generate disentangled features in its latent space and these features will then be used in the classification task. Once the classification task is complete, the relative importance of the features can be calculated and visualized via feature interpolation (changing the values) using the decoder component of the VAE. Hereafter, such features are referred to as "classifiable latent features."

The latent features are first separated into classifiable features, which are latent features of the VAE that are suitable for the classification, and non-classifiable features, which are the latent features of the VAE that are not suitable for the classification, in the latent space of the VAE. The most important classifiable features of an image can be identified after it has undergone classification. Hereinafter, the proposed model that can explain its decisions using explainable classifiable latent features (ECLF) is called an ECLF.

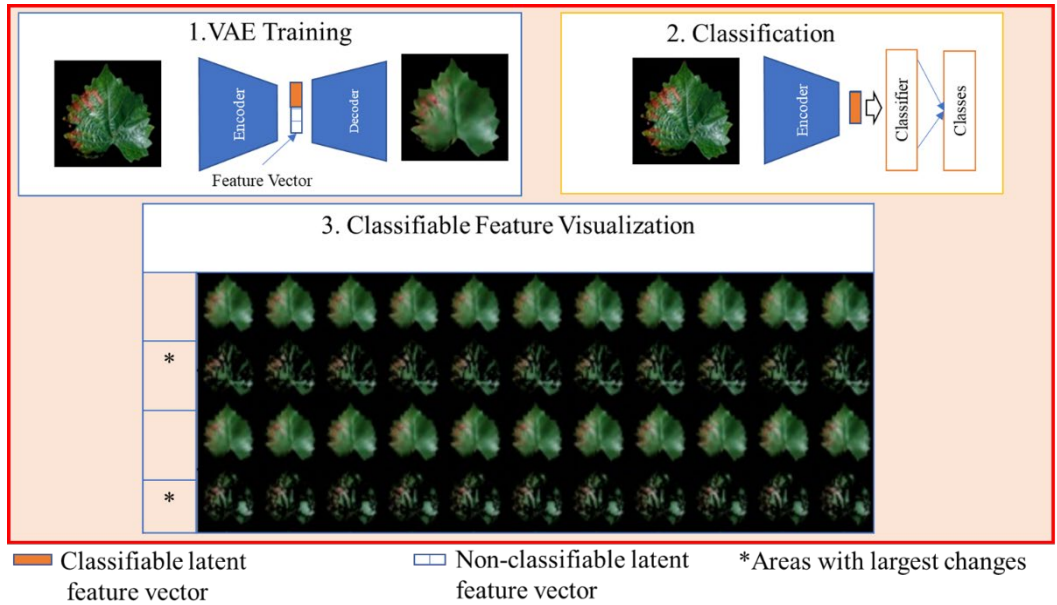

*Figure 1 Major stages of the ECLF system*

As shown in Figure 1, the latent features are distinguished as either classifiable and non-classifiable in stage 1. In stage 2, the classifiable features are used to train the image classifier (in this study, unless specified, the classifier refers to a nonlinear classifier). In stage 3, the variations of the features (left to right, from one class to another class) that are important to the classification are shown. In this image, the two most important features (1, 2) are shown. 1[*] and 2[*] show the changes in the features where the most significant changes occur. In addition, a complementary model, the explainable classifiable latent features class-specific (ECLF-CS) ~~system~~, is also established to extract class-specific latent features. Systems that apply these models are currently being developed (Oishi et al., 2021).

In the current study, the primary objective was to develop a system that can visualize the variation of important features in plant leaf images (diseased and healthy) and show the disentangled (separated variations) factors of these features that were used in the classification.

To confirm the explainability and accuracy of the proposed system, the explainability performance, the amount by which disentanglement affects the quality of the visual explanations and accuracy, and the major factors affecting disentanglement and explainability must be explored. To do so, the following objectives were set:

1. Investigate the visual understandability of,

   - classification decision explanations,

   - class differences

2. Investigate the effect of disentanglement (total correlation), which encourages the separation of features in latent space on,

   - decision explanation quality (quantitative) and classification accuracy

   - classification decision explanations (qualitative)

3. Investigate the effect of latent vector dimensionality on,

   - disentanglement (total correlation), explanation quality (quantitative) and

     classification accuracy

   - visual decision explanation quality (qualitative)

4. Test the ECLF-CS,

   - classification accuracy

   - explainability performance

5. Comparison of our methods accuracy with VGG-11 [29] accuracy

## 2.1. ECLF Model

The ECLF model is discussed in three major stages: 1. VAE training stage, 2. classifier training stage, and 3. Important feature visualization for classification.

### 2.1.1 VAE Training Stage

In this stage, the VAE is trained so that it learns to produce a disentangled latent feature vector that can extract the classifiable and non-classifiable features from an image and reconstruct that image. Three major factors were considered for building explainability in the VAE training stage:

1. Separation of classifiable and non-classifiable features in ECLF.

2. Disentangled representation for the separated features, which reduces the correlation.

3. Human understandability of the visualized features, which helps users identify important features from the reconstruction.

#### 2.1.1.1. Separation of Classifiable and Non-classifiable Features in ECLF

VAE is an algorithm that attempts to approximate a posterior distribution of the latent variable given a data point (Kingma & Welling, 2019), which is represented by an image. When an image is given to the encoder component of the VAE, it produces a latent vector ($z$) that is used as an input to the decoder component of the VAE. ECLF first divides the latent vector ($z$), which is produced by the encoder into two parts: a classifiable feature vector (*CFV*) and a non-classifiable feature vector (*NCFV*) using the following procedure (Figure 2) using an adversarial discriminator $f_d()$. If an image $\boldsymbol{x}$ is input into the encoder function $g(\boldsymbol{x})$, it produces parameters to a multivariate Gaussian distribution with diagonal variance ($\boldsymbol{\mu}$ - mean vector and $\boldsymbol{\sigma}^2$ - diagonal variance vector), where $i$ is the image index and $q_\phi$ is the posterior distribution of $\boldsymbol{z}$ given $\boldsymbol{x}$. $\boldsymbol{z}$ can be generated using Equation (1).

$$\boldsymbol{z} \sim q_\phi(\mathbf{z}|\mathbf{x}^i) = N(\mathbf{z}; \boldsymbol{\mu}^i, (\boldsymbol{\sigma}^2)^i \boldsymbol{I}), (1)$$

where $N$ is the normal distribution with a diagonal variance represented by $(\boldsymbol{\sigma}^2)^i \boldsymbol{I}$.

$$\boldsymbol{z} = [\boldsymbol{CFV}, \boldsymbol{NCFV}]. (2)$$

In this case, the $\boldsymbol{CFV}$ and $\boldsymbol{NCFV}$ parameters can be divided as follows:

$$\boldsymbol{CFV} = N(\mathbf{z}; \boldsymbol{\mu}^i_{cfv}, (\boldsymbol{\sigma}^2)^i_{cfv} \boldsymbol{I}), (3)$$

$$\boldsymbol{NCFV} = N(\mathbf{z}; \boldsymbol{\mu}^i_{ncfv}, (\boldsymbol{\sigma}^2)^i_{ncfv} \boldsymbol{I}), (4)$$

If the decoder function is $f(\boldsymbol{z})$, then the reconstructed image can be labeled as $\boldsymbol{x}'$ using Equation (5).

$$\boldsymbol{x}' = f([\boldsymbol{CFV}, \boldsymbol{NCFV}]) . (5)$$

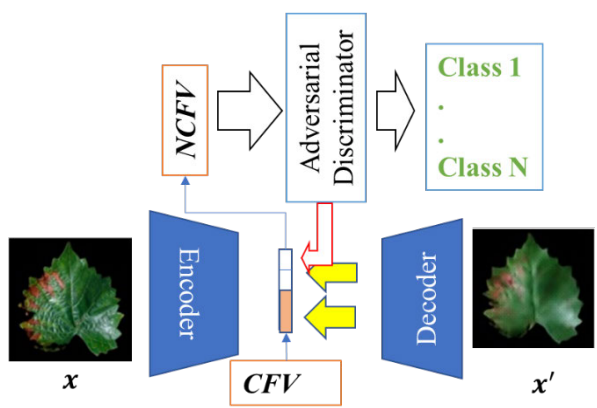

*Figure 2 Training procedure of the VAE and adversarial discriminator*

Figure 2 shows the simplified training procedure for the adversarial and variational autoencoders. Notably, an adversarial discriminator was used to form the **NCFV** during the VAE training. The red arrow indicates the discriminative loss from the adversarial discriminator. This loss (if the performance of $f_d()$ is good, it is considered as a loss) attempts to remove the classifiable features from the **NCFV**, and the yellow arrow shows the reconstruction loss from the decoder to encoder.

An adversarial discriminator function $f_d()$ attempts to learn the class of the input image using only **NCFV** [26]; therefore, $C_d$ is the class assigned to $x$ by $f_d()$ using **NCFV**.

$$C_d = f_d(NCFV), (6)$$

Given that the actual class of $x$ is $C_{gt}$ (the ground truth class), $f_d()$ is trained using the classification loss between $C_{gt}$ and $C_d$. In contrast to (Lample et al., 2017), which used an autoencoder, a part of the VAE output was used, and the **CFV** was not conditioned using an attribute label because the algorithm should find the classifiable attributes or features by itself. If the adversarial discriminator can learn to discriminate the classes with a desirable accuracy, then, the **NCFV** might contain information that can be used to clearly separate the classes. Given that adversarial training is conducted, if the accuracy is high in $C_d$, the prediction is considered as a loss $\mathcal{L}_d$ for the encoder (Equation (7)). Therefore, the decoder is discouraged from producing classifiable features in the **NCFV**.

$$\mathcal{L}_d = lossfunction(C_d, \neg C_{gt}). (7)$$

Gradually, the encoder learns to send classifiable features through the **CFV**. Therefore, the VAE loss function was minimized while also attempting to minimize $\mathcal{L}_d$.

A supportive classifier for the **CFV** was adopted to support the formation of classifiable features. The classifier plays a supportive role in the system and is trained in parallel to the discriminative classifier. It performs an opposite function to the adversarial classifier (Figure 2) by providing feedback to the VAE on the classifiability of the features.

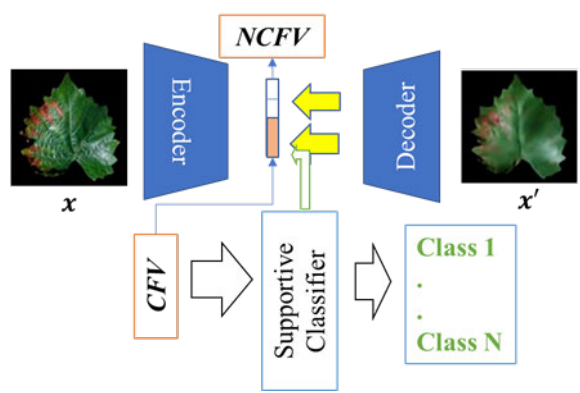

*Figure 3 Role of supportive classifier in the VAE training*

Figure 3 shows the manner in which the supportive classifier is used in the training of the VAE, in contrast to the red arrow shown in Figure 2. In Figure 3, the loss (green arrow) is created such that the features produced in the **CFV** support the classification in the supportive classifier. However, this loss was set such that its role is not as prominent as compared to the adversarial classifier. Although this classifier can be used for the final classification (stage 2 in Figure 1), the same hyperparameters of this classifier were during the VAE training; this leads to suboptimal classification accuracy. For this reason, a separate classifier was used as the final classifier.

A convolutional encoder and decoder were used in the system architecture. Note that it is critical to ensure that the features extracted in the encoder are properly represented in the decoder. To achieve this, a decoder with the same weights as the encoder was constructed. The authors of (Arora et al., 2015) proved that hidden layer activations in a DNN can be recovered using a generative network. As such, it was assumed that the VAE encoder can be recovered using the decoder if proper conditions can be ensured. The restrictive nature of the VAE loss function, which has an information bottleneck property (Alemi et al., 2017), (Burgess et al., 2018), may prevent **NCFV** features from going through the **CFV**.

### 2.1.1.2. Latent Vector Disentanglement in VAE

Disentanglement in the latent feature vector $z$ is encouraged. In **CFV**, if:

$$[cf_1, cf_2, cf_3 \ldots \ldots \ldots \ldots \ldots cf_n] = \boldsymbol{CFV}, (8)$$

Then, when the latent vector is decoded while changing one **CFV** feature $cf_n$, only the image features that correspond to $cf_n$ should change in $x^{'}$. This is a necessary condition when visualizing which latent feature corresponds to which image features in $x'$. Although the definition of disentanglement has been a debated topic (Mathieu et al., 2019) in recent years, several methods have been proposed to help variational autoencoders learn disentangled latent features (Burgess et al., 2018; Higgins et al., 2016; Kim & Mnih, 2018). The current study used the algorithm presented in (T. Q. Chen et al., 2018) to improve the disentanglement between the features during training (Equation (9)) owing to its ability to factorize the different factors of the VAE loss. Thus, only the required factors were minimized.

$$\mathcal{L}_\beta = \frac{1}{N}\sum_{n=1}^{N}\big(\mathbb{E}_q[\log p(x_n|z)] - \beta \ KL \ (q(z|x_n)||p(z)). \ (9)$$

An attempt was made to increase the evidence lower bound, which is $\mathcal{L}_\beta$, by reducing the second term of the $KL$ divergence between the two probability distributions $q(z|x_n)$ and $p(z)$ of Equation (9), which can be considered as the information bottleneck term (Alemi et al., 2017), (Burgess et al., 2018). Notably, $N$ denotes the number of images in a batch, and $\beta$ denotes the coefficient responsible for controlling the information bottleneck.

According to (T. Q. Chen et al., 2018), the second term can be divided into three components:

$$index \ Code \ MI \ = \ KL \ (q(z,n)||q(z)p(n)), (10)$$

$$Total \ Correlation \ = \ KL \ (q(z)||\textstyle\prod_j q(z_j)), (11)$$

$$Dimension \ wise \ KL = \textstyle\sum_j KL(q(z_j)||p(z_j)), (12)$$

$$\mathbb{E}_{p(n)}[KL \ (q(z|n)||p(z))] = \ index \ Code \ MI + \ Total \ Correlation(\boldsymbol{TC}) +$$

$$Dimension \ wise \ KL(\boldsymbol{DKL}). (13)$$

Of these terms, the total correlation term in Equation (11), which reduces correlation and encourages independence, and dimension wise KL in Equation (12) were given special attention.

### 2.1.1.3. Full Training Loss Function for the Feature Separation Stage

For the training, the total loss function expressed in Equation 14 was used. In this function, the reconstruction loss was denoted by $L_{rc}$, the training loss of the VAE by $L_{VAE}$, the supportive classifier loss by $L_s$, the adversarial discriminator loss by $L_d$, and the discriminative regularization loss (Lamb et al., 2016), which is used to increase the visualization quality, by $L_{rd}$, and $\alpha, \epsilon, \varepsilon, \beta$ and $\gamma$ were used as weights during the training.

$$L_{VAE} = L_{rc} + \alpha L_{rd} + \epsilon L_d + \varepsilon L_s + \beta TC + \gamma DKL. \text{ (14)}$$

In this function, the losses can be divided into four main categories; the $L_{rc}$ and $L_{rd}$ can be referred to as the reconstruction losses, which are helpful in identifying the features reconstructed or changed by changing the $cf_n$ of the VAE; $L_d$ and $L_s$ can be referred to as the feature separation classifiable losses, and $\boldsymbol{TC}$ can be referred to as the disentangling loss, and $\boldsymbol{DKL}$ as the prior matching loss.

### 2.1.2. Classifier Training Stage

The classifier is responsible for recognizing classes using $\boldsymbol{CFV}$. After the VAE training was completed, the classifier was trained on the $\boldsymbol{CFV}$. Given that the encoder produces parameters for the distribution, $\boldsymbol{CFV}$, as shown in Equation (3), the classifier cannot be trained on this distribution. Since this might reduce the classification accuracy, the value that has the highest likelihood to appear, which is $\boldsymbol{\mu}_{cfv}^i$, was selected to train the classifier while keeping the encoder weights fixed. After the classifier is trained, it can be used to make predictions (final classification), for which $\boldsymbol{\mu}_{cfv}^i$ was used as the input to the classifier. Once the prediction is made, it needs to be explained using the features learned in the VAE. If a linear classifier is used, it is easy to understand which features are more important than the others because the relationship between the features and the prediction is linear. However, for a more generalized approach, it is better to use a classifier algorithm that can use both nonlinear and linear relationships for classification. Several types of nonlinear classifiers can be used in this part of the algorithm.

### 2.1.3. Important Feature Visualization for Classification

Through this visualization, which features play an important role in selecting one class from another are determined. When the nonlinear classifier makes a decision on an image, it uses a local decision boundary to decide between the classes; on a zoomed-in level, this decision boundary can be approximated using a linear classifier. This method (local interpretable model-agnostic explanation; LIME) was introduced in (Ribeiro et al., 2016) and uses a standard segmentation algorithm to create image super pixels (sets of pixels with the same properties), then masks other super pixels to make sample points (input point to the algorithm). Numerous methods to determine the importance of a feature in decision-making are available. However, because local decision boundaries were the focal point of this study, we used an approach similar to the LIME method to determine the importance of a feature.

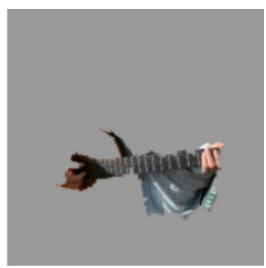

*Figure 4 LIME explaining electric guitar detection* (Ribeiro et al., 2016)

Figure 4 shows that the LIME algorithm determines the pixels responsible for the selection of the acoustic guitar class; however, LIME cannot show feature variations. <u>In the proposed method, the VAE was trained to show the variations of the classifiable features</u>. This led to a model-dependent explanation owing to the process by which the sample points were created.

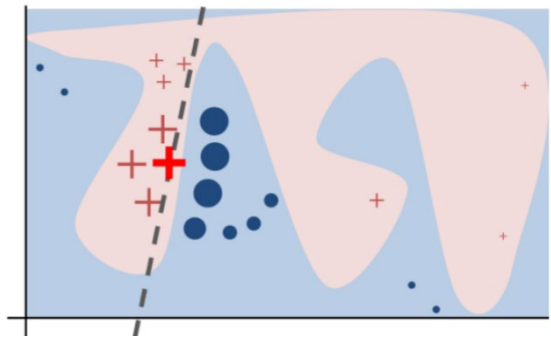

*Figure 5 LIME toy example* (Ribeiro et al., 2016)

In Figure 5, the blue and pink backgrounds are separated by the nonlinear function of the deep learning model. This nonlinear function cannot be approximated using a linear function; LIME attempts to explain the bright red cross point by sampling instances (data points) and sending them to the nonlinear function to obtain the predictions; they are then weighed according to the proximity, which is denoted by the size of the marker. A faithful local explanation is denoted by a dashed line (Ribeiro et al., 2016).

### 2.1.3.1. Approximating Nonlinear Boundary Using a Liner Classifier

The proposed method focuses on two classes that are used to explain a decision, namely; class A and class B. <u>Class A is selected by the nonlinear classifier as the maximum likelihood class for the *CFV* that is being explained (this can be different from the $C_{gt}$)</u>, meaning that the features that differentiate class A from class B need to be known.

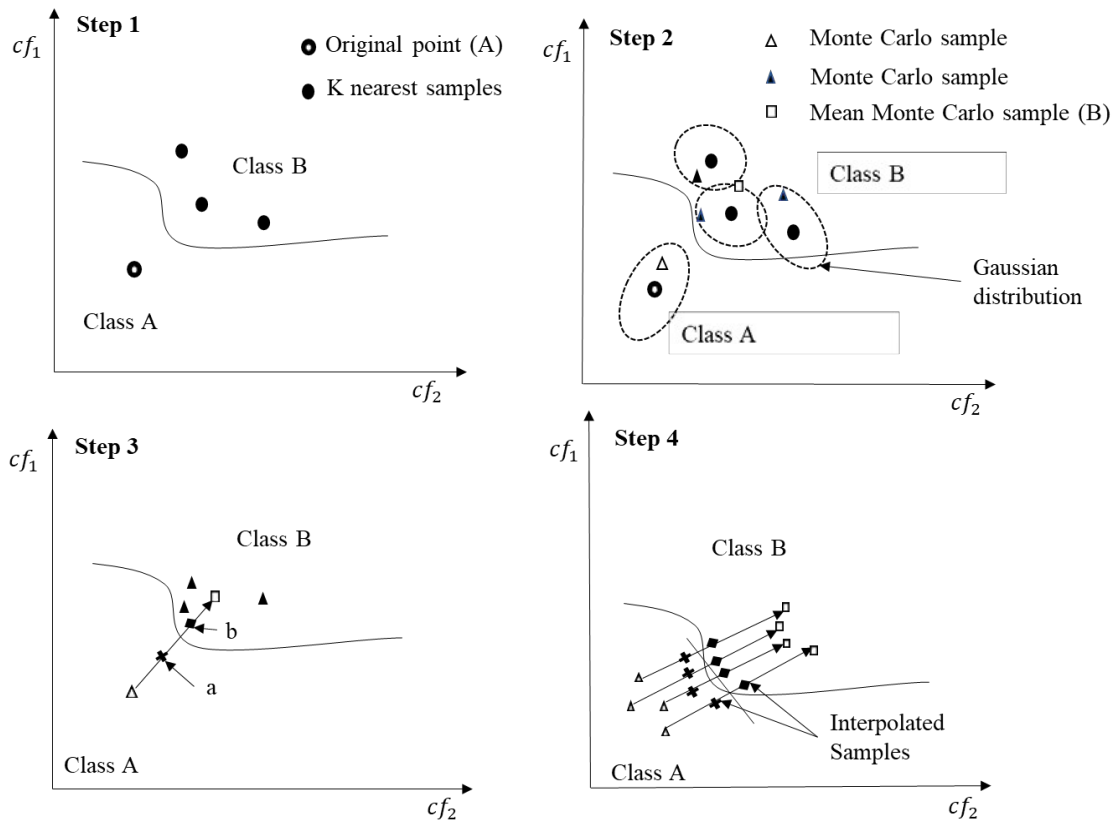

*Figure 6 Data Point Creation for linear classifier training*

In Figure 6, Step 1 shows the original point produced by class <u>A image's **CFV** that needs to be explained. Then, **CFV** samples must be identified in class B using the K-nearest samples to the original point.</u> In Step 2, a new point from point (A) is obtained via Monte Carlo sampling using $\mu_{cfv}^i, \sigma^2{}_{cfv}^i$ of the original point. Monte Carlo samples are produced with a Gaussian distribution for each class B sample, as shown in Step 1, and then averaged to obtain point (B). In Step 3, an interpolation from point (A) to point (B) is determined while testing if the interpolated point has crossed the decision boundary. Via this procedure, points (Explanation support points) (a) and (b) that are the nearest to the nonlinear decision boundary and belong to classes A and B, respectively, are obtained. Steps 2 and 3 are repeated to obtain several point (a)s and point (b)s. In Step 4, these points are used to train a linear classifier.

In Step 2, the samples are taken from the distributions of the samples because it represents the probability distribution of that point. Using these points, the linear plane perpendicular to the variation of the variable that contributes the most to the classification can be constructed.

The linear classifier was trained to approximate the activation of the nonlinear classifier in the sampled region.

### 2.1.3.2. Selecting Most Important Features for Classification

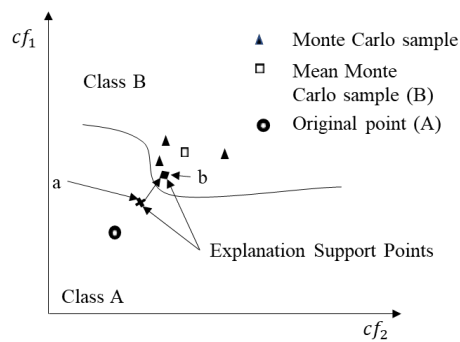

*Figure 7 Explanation support points*

Points (a) and (b) were chosen to be closer to the decision boundary, as shown in Figure 7; after this, the importance was defined as the cause of the change during the activation of the linear classifier for class A when $CFV_a$ and $CFV_b$ were taken as inputs The following formula was used to determine the importance of each feature: if $W$ is the weight vector, and

$IM$ is the importance, then

$$IM = W_A\,(CFV_a - CFV_b),\,(15)$$

The maximum likelihood class was chosen over $C_{gt}$ because the nonlinear classifier sometimes yields incorrect classifications. In such a case, an explanation must be given as to why such a decision was made by the nonlinear classifier.

### 2.1.3.3. Visualizing Most Important Feature for Classification

Once the most crucial features were selected; the meaning of these features and their variations must be known. To understand this, the important features were selected and feature interpolation was performed, where a given feature value of the features of interest was changed toward the other class while keeping the other feature values constant. The change that corresponds to a particular image and classifiable features in the reconstructed image $x^{'}$ was visualized by combining Equations (5) and (8).

$$x^{'} = f([[cf_1, cf_2, cf_3\,\ldots\ldots cf_n\,\ldots\ldots],NCFV]),\,(16)$$

$$\triangle x^{'} = \frac{\partial f}{\partial cf_x}.\,(17)$$

When a classifiable feature is visualized to exactly understand that feature, the way that the features respond to the change in the latent variable $cf_n$ also needs to be visualized by varying $cf_n$ and visualizing the changes in $x^{'}$. This is shown in Eq. (17):

The interpolation was performed by changing each classifiable feature or set of classifiable features from the supportive point of class that needed explanation for the other class (point (a) toward point (b) or from point (A) toward point (B)). There are two ways to visualize this: single-factor-wise interpolation or factor group interpolation. For the factor-wise method, Figure 8 shows how the original class point from class A is changed to the mean point in class B.

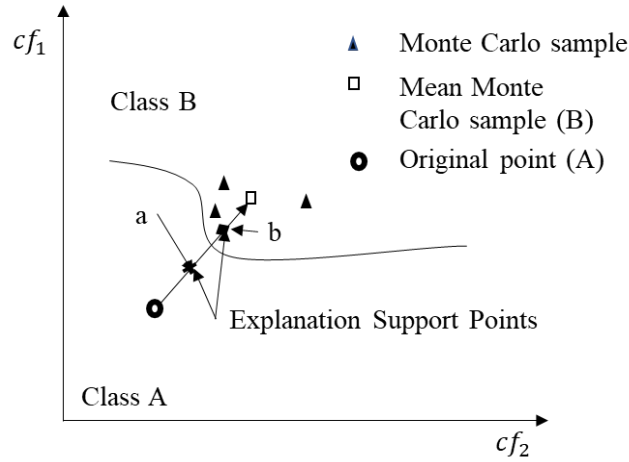

*Figure 8 Feature interpolation for visualization*

To visualize how the features change in the local area where the decision is made, the proposed method creates a representative data point by averaging the K-nearest points of class B, as shown in Figure 8. This point is referred to as the mean K-nearest point. The individual feature value $cf_n$ is changed from point (A) to point (B) to visualize how an individual feature change affects $x'$. Although this is not an ideal condition, a feature interpolation starting before points (a) and (b) may need to be performed to enhance the visualization of the feature.

Equation (19) shows how an interpolation vector is created:

$$cf_n^{AI} = cf_n^A - (cf_n^A - cf_n^{BM}) \times k. \text{ (18)}$$

Where $cf_n^{AI}$ is the changed input feature for the visualization, $cf_n^A$ is the feature value of class A at the original point, and $cf_n^{BM}$ is the mean point feature value of class B; $n$ is the feature number for the visualized feature, and $k$ is an interpolation constant.

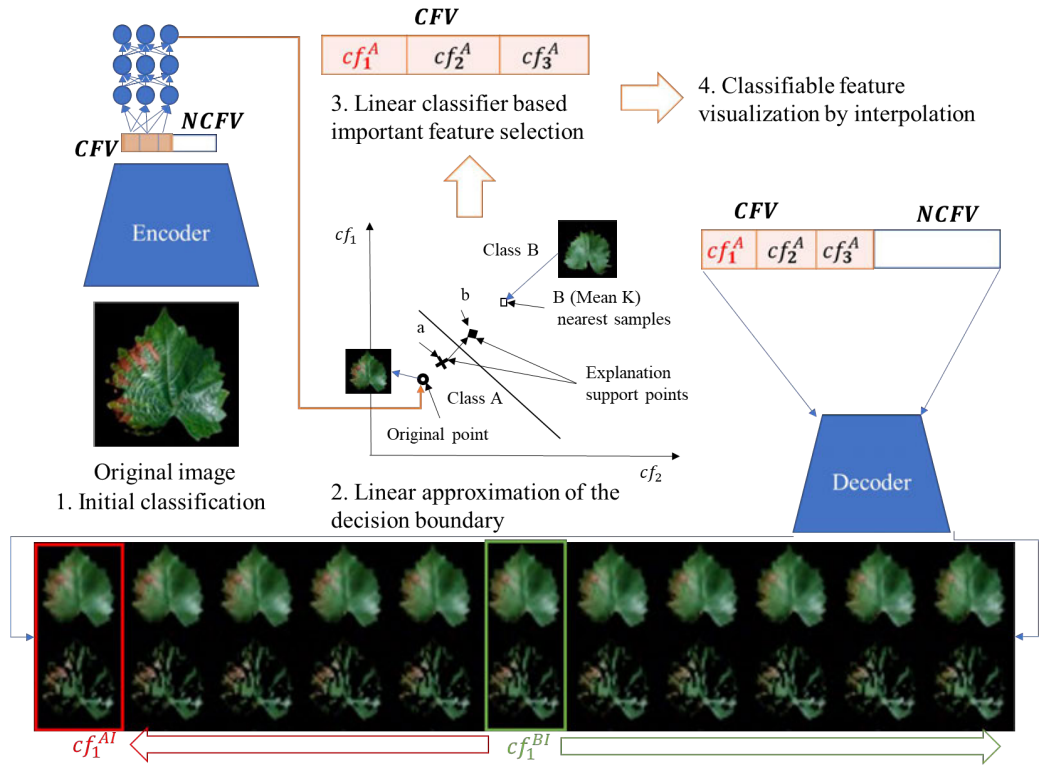

*Figure 9 Classification explanation overview*

Figure 9 shows the procedure for conducting important feature selection and visualization in the ECLF system: <u>first, an initial classification is made, second, the linear classifier makes an approximation of the nonlinear decision boundary, and third, the most important features are selected based on linear approximation. Finally, the most important features are selected, for example, $cf_1^A$ in this instance, is visualized by interpolating from point (A) to point (B) and toward point (B),</u> which is shown in Equation (18). The bottom row shows the results of the interpolation: the variations of the most important feature that contributed to the classification decision.

The next method involves selecting a group of features and then changing them, for example, the top 10 features that are important for the classification. This method can be used when the changes in individual features are not very prominent. This method can also be extended to visualize all the classifiable features.

### 2.1.3.4. Visualization of Areas with the Largest Change

Although it is sometimes possible to visualize the changes in $x^{'}$, the exact areas where the features have changed can be easily visualized.

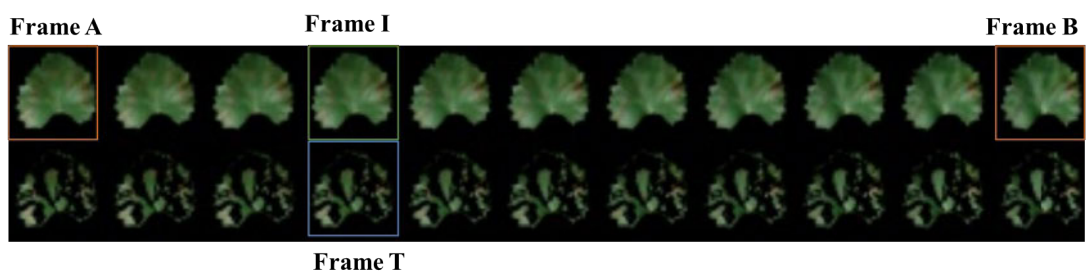

*Figure 10 Visualizing the Changed Areas*

In Figure 10, to visualize frame T, the absolute differences in frame I as we move from frame A to B as well as the absolute differences across the channel were added. Then, the $80^{th}$ percentile of the added values, and the threshold of each channel of frame T according to the remaining values were considered to capture the differences between frames A and B.

### 2.1.4. Factors Effecting Explainability and Accuracy in ECLF

Explainability and accuracy play major roles in any explainable system. Many researchers who work with explainable algorithms have expressed concerns regarding the tradeoff between explainability and accuracy (Luo et al., 2019; Veiber et al., 2020). Moreover, explainability has several definitions in the literature on deep learning (Xie et al., 2020). In the ECLF system, the definition of explainability and accuracy needs to be clarified and the important factors that affect these two properties and their tradeoff needs to be determined. As discussed in Section 2.1.1.2., during training, the value of $\beta$ increases the loss to the disentanglement term, helping to reduce the disentanglement. This means that the disentanglement term is dependent on the value of $\beta$.

In the current study, different aspects of explainability were considered, such as the interpretability of the classifiable features (disentanglement, compactness, and separation of the **CFV** and **NCFV**) and human understandability (visual quality) of the produced explanations. In Section 2.1.1.1., the importance of disentanglement for explainability is detailed. When the compactness is considered, the lower dimensionality of the **z** increases the compactness wise versa; so, the user must go through a low number of dimensions to investigate, increasing the explainability. Moreover, separating the **CFV** and **NCFV** increases the explainability of the system as it helps us determine which features are classifiable and which are not. Users can understand each feature by visualizing the changes in $x^{\prime}$ (Section 2.1.3), therefore, the quality of $x^{\prime}$ also plays a paramount role in the explainability of ECLF. The interaction between these factors needs to be understood to obtain a better understanding of the explainability of the ECLF system, more specifically, how increasing $\beta$ (Section 2.1.1.2), which scales the $TC$ term of the loss function, would affect the other explainability components. In addition, the dimensionality of $z$, which might play a paramount role in explainability, also needs to be investigated.

The final classification accuracy of the trained nonlinear classifier during the classifier training stage is considered as the overall accuracy. Given that the input for the final classifier is $CFV$, which is affected by disentanglement ($\beta$ in this case) and the dimensionality of $z$, the interaction between the classification accuracy and the explainability factors such as disentanglement and dimensionality also needs to be understood.

## 2.2. Explainable Classifiable Latent Features Class-Specific (ECLF-CS)

Even though the ECLF model can show the differences between <u>classifiable features of individual instances of classes, it cannot show class-specific features.</u> To avoid this complication, a system was developed that can show the class-specific features of a $CFV$. <u>A class-specific feature vector can be defined as a feature vector that is specific to one class and can separate that class from at least one other class.</u> Although it is possible to use ECLF-CS in multiclass situations, this investigation only considered two-class situations in this study. ECLF-CS uses a latent vector that contains classifiable features specific to classes in two separate vectors, called $CFVS_1$ and $CFVS_2$, which in turn are assigned to two classes, called $c_1$ and $c_2$. In this case, the latent vector $z$ can be expressed as

$$z = [CFVS_1, NCFV, CFVS_2]. \text{ (19)}$$

In the encoding phase, an input image $x$ is sent through the encoder to produce the $z$ vector and the input to the decoder is decided based on the class of $x$. Only the vector assigned to that class and the $NCFV$ are passed to the decoder; for example, if the class is $c_1$, the vector passed to the decoder is $[CFVS_1, NCFV]$. For the $NCFV$, the adversarial loss is used, as in Equation (7). <u>The number of classes that can be trained is limited in this approach, and training is slow compared to ECLF because of the class-specific training procedure.</u>

### 2.2.1. Differences between ECLF and ECLF-CS

As explained in Section 2.2, ECLF and ECLF-CS features have different characteristics; ECLF feature present the characteristics of the entire dataset, while ECLF-CS features only present the classifiable characteristics of a single category of features. Therefore, ECLF-CS features can provide direct information on the presence of a given category.

### 2.2.2. Classification and Important Feature Visualization for ECLF-CS Features

The classifiers for ECLF-CS are trained in the same manner as the classifier for ECLF. Since there are two class-specific vectors ($CFVS_1$ and $CFVS_2$), they are merged before they are trained.

$$CFVS = [CFVS_1, CFVS_2]. \quad (20)$$

The important features were determined and visualized according to the procedure described in Section 2.1.3. In contrast to ECLF, when a feature is selected as important in ECLF-CS, feature visualization is performed depending on the feature vector from which the feature is obtained. If it came from $CFVS_1$, it is possible that $CFVS_2$ was not used to produce $x^{'}$ and vice versa.

## 2.3. Technical Details on Architecture and Training

### 2.3.1. VAE Architecture

During training, the convolutional layers were pretrained using the entire PlantVillage

dataset (Hughes & Salathe, 2015), (Mohanty et al., 2016), which contains diseased and healthy

plant leaves. Then, training on specific datasets was conducted, which is discussed in Section

2.4.

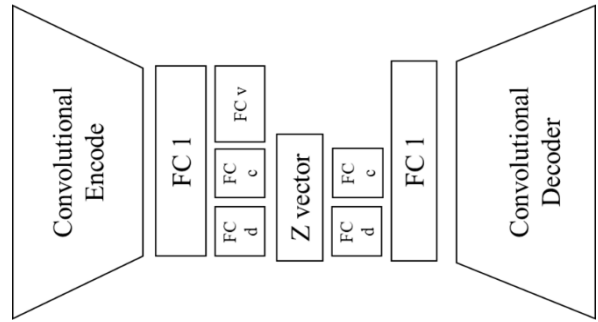

*Figure 11 Network architecture of the VAE*

As shown in Figure 11, the network architecture of the VAE convolutional network

has five layers. Fully connected layer FC1 had 8,192 inputs and outputs, fully connected layers

FC-d and FC-c, which produced the **CFV** and **NCFV**, each with an input of 2,048. The fully connected layer FC-v, which produced the log value of the variation ($\log(\boldsymbol{\sigma}^2)$) for the variation of variables in the **CFV** and **NCFV** (discussed in Section 2.1.1) had a size of 4,098, and the output size of the fully connected networks was determined by the latent vector. The input and output sizes of the network are $128 \times 128 \times 3$.

### 2.3.2. Architecture of the Discriminator and Supportive Classifier

All the classifiers and discriminators have three fully connected layers with a rectified linear unit (ReLU) and the output size of the ***CFV*** determines the input size of the supportive classifier.

### 2.3.3. VAE Training Stage

The VAE is pretrained using the entire PlantVillage dataset for 106,000 iterations, using only the convolutional parts, which together act as a pure encoder–decoder architecture.

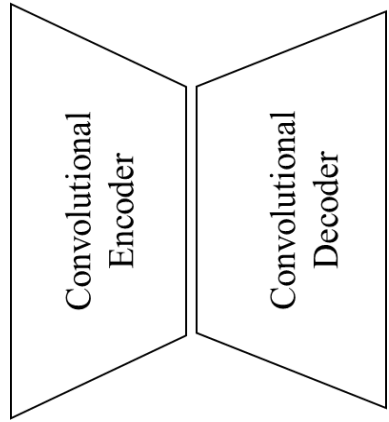

*Figure 12 Encoder–decoder architecture for pretraining*

Up to 120,000 iterations, the system was trained using $L_{rc} + \epsilon L_d + \varepsilon L_s$ as shown in Equation (14). The warmup phase (Sønderby et al., 2016) was then used on $L_{rd}, TC$, and $DKL$ for training up to 140,000 iterations. Although the warmup phase used KL divergence terms from, the warmup phase on $L_{rd}$ was also used to balance the learning with the KL divergence terms. Although the results were saved and calculated every 20,000 iterations, the final results were considered after 1,500,000 iterations.

Ideally, $x^{'}$ would be a good representation of $x$; however, this is not always possible and in this case, the faithfulness of the reconstruction comes into question. In recent years, several researchers, including (Lamb et al., 2016), have attempted to solve this problem. Following (Lamb et al., 2016), discriminative regularization loss, namely a VGG-16 network [29], was used as a discriminative regularizer, and the first three layers of the network were used for discriminative regularization.

### 2.3.4. Final Classifier Training Stage

The training was conducted for 5,000 iterations and the best validation accuracy was used for the testing. The classifier in the final classifier training had the same architecture as the supportive classifier.

### 2.3.5. Explanation Generation and Visualization Stage

#### 2.3.5.1. Linear Classifier Training

The linear classifier, which is used in the explanation phase, was trained using samples generated by the VAE, as explained in Section 2.1.3. The linear classifier input was the same as the size of the *CFV*, and the linear classifier had two outputs. One thousand sample points were used to train the linear classifier and the mean squared error (MSE) between the nonlinear classifier activation and linear classifier activation was minimized to represent decisions near the boundary.

## 2.4. Experimental Data

Part of the PlantVillage dataset from (Hughes & Salathe, 2015), (Mohanty et al., 2016),

which contained 39 classes of images (12 healthy and 27 diseased classes), which contained a single leaf in each image, was used in this study. Segmented versions of the leaf images, which were created in (Mohanty et al., 2016) were used.

If the full PlantVillage dataset is used, the system would have to learn not only the differences between the respective diseases and healthy and diseased plants but also the differences between the plant types, for example, the differences between grape and potato leaves. In a real field or application condition, however, the type of plant is known and only the distinction between healthy and diseased leaves needs to be determined. Therefore, original datasets that contained only one type of leaf using the PlantVillage dataset (Table 1) were created for this purpose. Since some classes like "potato healthy" had very few images, the datasets needed to be restricted to 30 validation and testing images per class.

*Table 1 Dataset Statistics*

| Original Dataset | Containing Classes | Number of Training Samples | Number of Validating Samples/Class | Number of Testing Samples/Class |
|---|---|---|---|---|
| **Grape4** | Healthy, Black Rot, Black Measles, Leaf Blight | 3,823 | 30 | 30 |
| **Grape2** | Healthy, Leaf Blight | 1,379 | 30 | 30 |
| **Apple4** | Healthy, Black Rot, Rust, Scab | 2,931 | 30 | 30 |
| **Apple2** | Healthy, Scab | 2,155 | 30 | 30 |
| **Potato3** | Healthy, Early | 1,972 | 30 | 30 |

| | Blight, Late Blight | | | |
| --- | --- | --- | --- | --- |
| **Potato2** | Healthy, Early Blight | 1,032 | 30 | 30 |

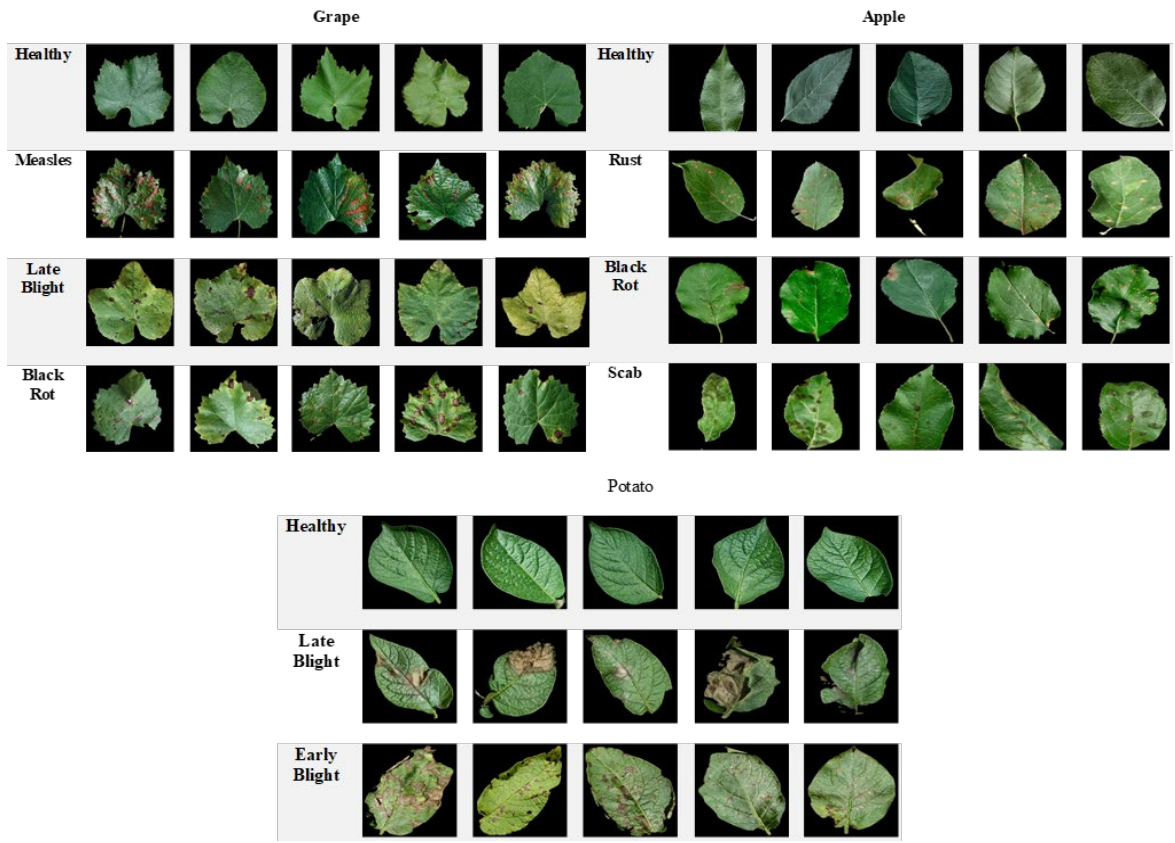

*Figure 13 Sample leaves from three original datasets*

Figure 13 shows sample leaves from different datasets as well as features that were used to differentiate the classes. The apple dataset seems to have more subtle features compared with the other two classes.

## 3. Experimental Results and Discussions

### 3.1. Visualization of Decision Explanations of ECLF

The 320-dimensional latent vector VAE was trained for 1,500,000 iterations, which increased the classification accuracy and the reconstruction loss. For the Grape4, Apple4, and Potato3 datasets, the classification accuracies were 96.7, 90.0, and 94.4%, respectively. The visual quality of the reconstructions increased, although the reconstruction error seemed to increase owing to overfitting. Therefore, it was assumed that the classifiable features require more iterations for training.

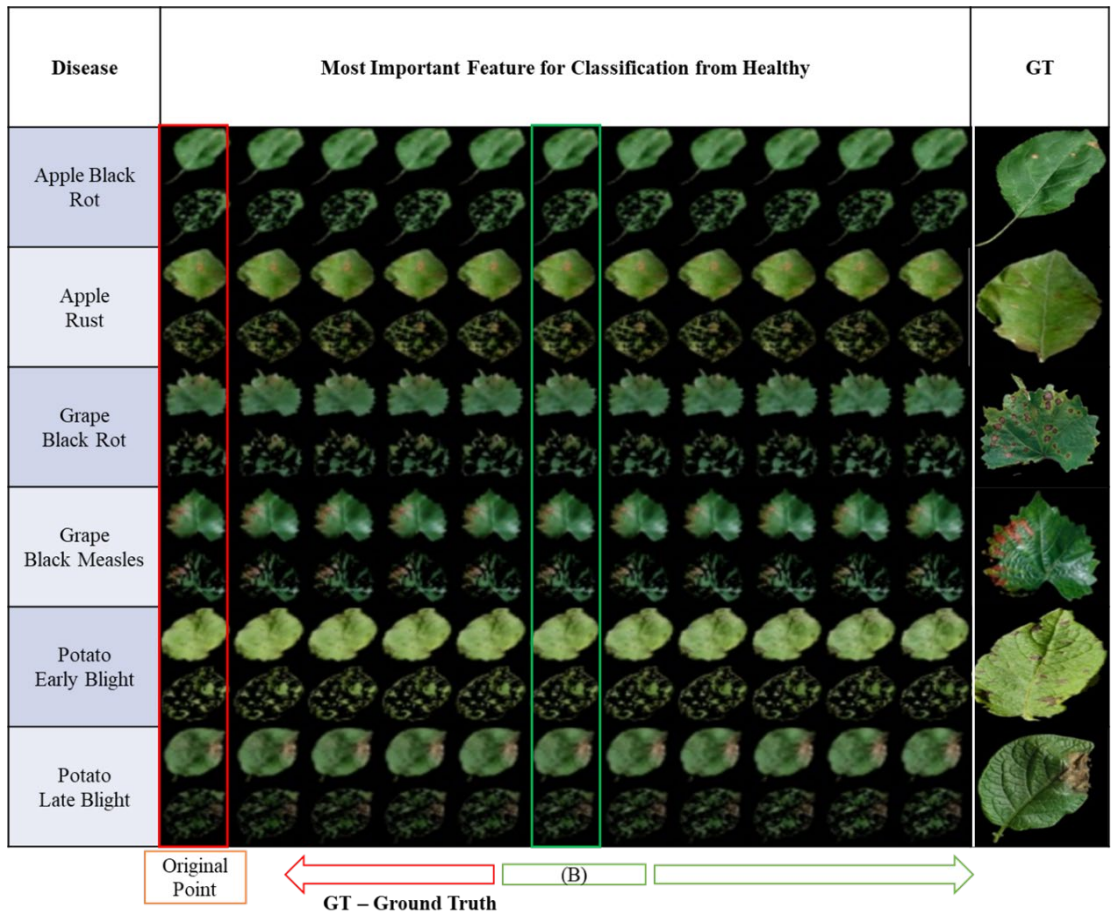

**A**

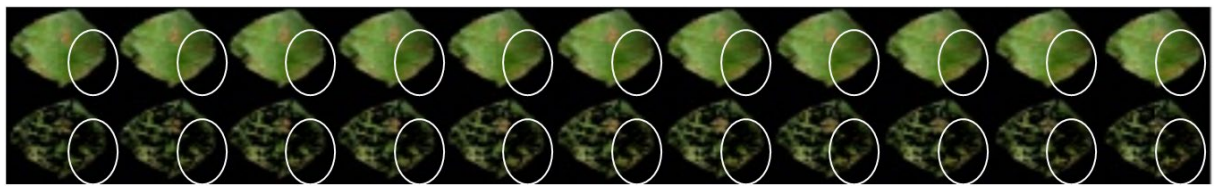

**B**

*Figure 14 **A** - Sample leaves from 3 datasets, **B** - Feature encroachment to healthy side*

Figure 14 **A** depicts the approach via which the features that are the most important for the classification gradually shift from the diseased class to healthy class; features in the real image are represented in the original point column. It can also be seen that a combination of shape and color changes was considered as a single latent feature by the VAE. Almost all the images showed an increase in the green color areas when the features shifted from the diseased column to the healthy column. Some images showed a reduction in the number of brown patches. These features were particularly prominent in grape, measles, and black rot when the features shifted further from the original point to the mean healthy point and beyond.

Figure 14 **B** depicts the encroaching of other diseased class features on the healthy class features. This phenomenon (encircled with a white ellipse) is marked in the bottom right side of the apple rust interpolation and may have occurred because this feature represents the variations of the classifiable features across all the classes. Thus, the feature from Figure 14 **B** may help separate the diseased and healthy classes locally but can contain other class features. Several difficulties were encountered when obtaining the explanations, with one of the most severe being that the reconstruction was not 100% accurate. The decoder recreated the latent vector supplied by the encoder, and some information was subsequently lost in the latent vector, which may have led us to this loss in reconstruction accuracy. Research is being conducted to create a VAE that can generate images that are very close to the original image (Hou et al.,

2019). Moreover, sometimes, the explanations are not easily detectable by the human eye.

### 3.2. Visualization of Class Differences

Traveling from point (A) to point (B) in Figure 15, the manner by which the classifiable features change from class A to class B can be visualized. Given that it is easy to interpret such changes between diseased and healthy classes, the diseased class was used as class A and the healthy class as class B to visualize the features changing from class A to class B. Moreover, it is also advantageous to understand what features were used close to the decision boundary. Therefore, travel was also conducted from point (a) to point (b) for visualization purposes.

The 10 most important features for interpolation were visualized while keeping other features constant.

Figure 15 shows how the classifiable features change from diseased to heathy when traveling from point (A) to point (B) and from point (a) to point (b) in the classifiable feature space.

In this stage, users can identify the features that collectively change from diseased to healthy. Brown patches, yellowish color, and the degree of leaf damage probably affect the classification between diseased and healthy classes. Users can obtain an idea of these changes by observing the changes that took place from point (A) to point (B); however, the decision is made using the changes that occur near the decision boundary, from point (a) to point (b). These features were also the ones that showed the largest change near the decision boundary, and although the features are not very prominent, they can be identified with the help of visualization from point (A) to (B).

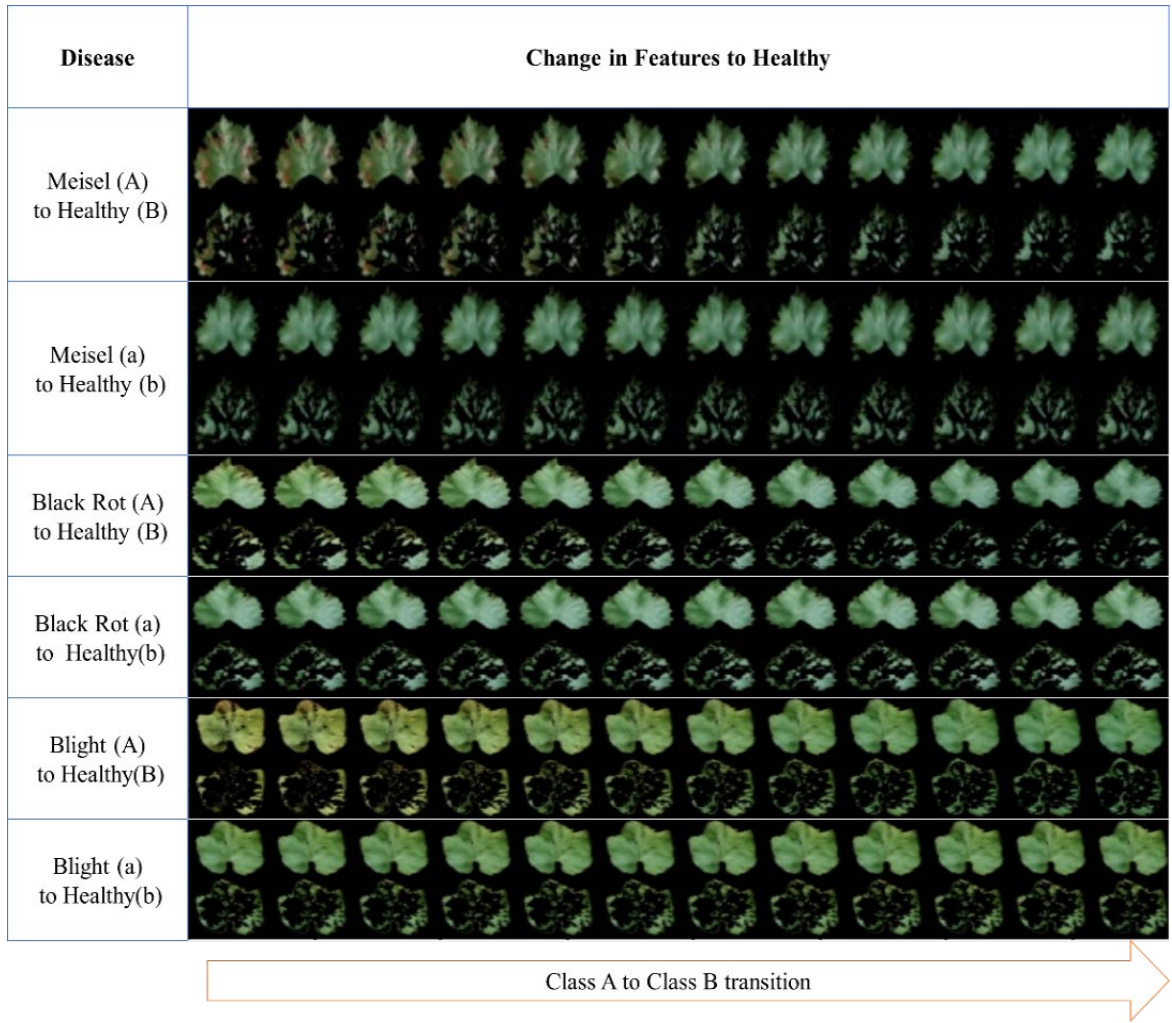

*Figure 15 Change of Features from Diseased to Healthy*

### 3.3. Effect of Total Correlation Constant $\beta$ on Total Correlation, Classification Accuracy, and Reconstruction Loss of ECLF

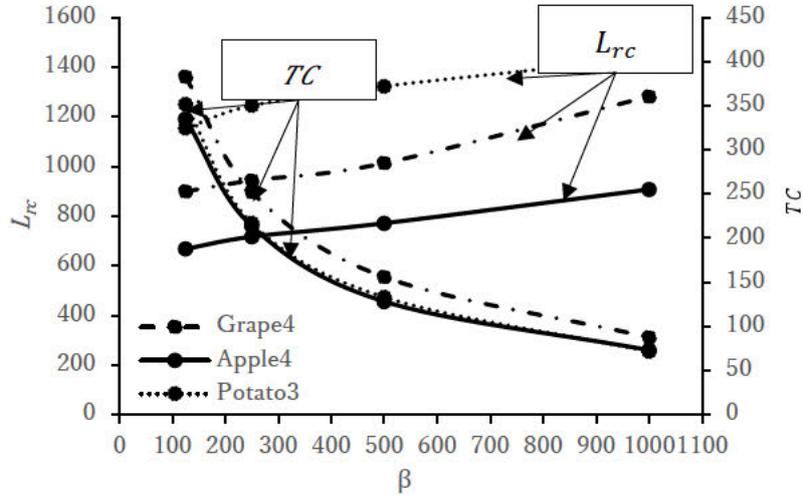

*Figure 16 Effect of $\beta$ on $L_{rc}$ and $TC$*

Testing was conducted using 320 latent dimensions of the VAE and $\beta$ was changed to 125, 250, 500, and 1,000 while keeping $\gamma$ constant.

Figure 16 shows the effect of $\beta$ on the $TC$ and $L_{rc}$ values, where $L_{rc}$ appears to increase linearly. Considering that $TC$ has a linear relationship with the dimensionality in Figure 16, the exponentially higher values of $\beta$ must be used to reduce $TC$ linearly.

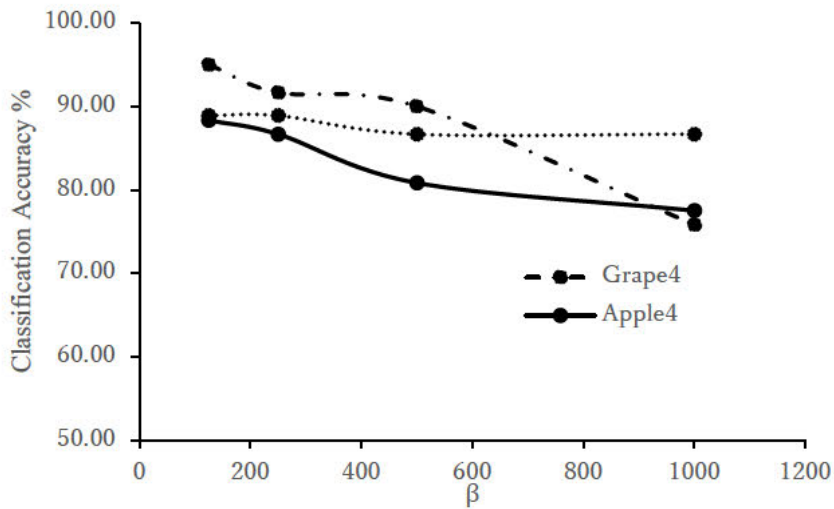

*Figure 17 Effect of $\beta$ on classification accuracy*

Figure 17 shows the manner in which the classification accuracy changed with $\beta$. As can be seen, the classification accuracy decreases on the Grape4 and Apple4 datasets and in Figure 18, <u>the information of high-frequency features is reduced with a higher $\beta$ value.</u> This was attributed to the Grape4 and Apple4 classes depending on high-frequency information, such as spots and lesions, in contrast to low-frequency information such as color. <u>For the Potato3 classes, the effect of increasing $\beta$ on classification accuracy seemed to be low when compared to the other two classes, which might indicate that the classifiable features belong to the low-frequency information of the Potato3 images.</u>

### 3.4. Effect of $\beta$ on the Decision Explanation of ECLF

Figure 18 shows the effect of $\beta$ on the 320-dimensional VAE and shows the top two most important features for each dimension for the visualization of the discrimination between grape late blight class and healthy class. From the top of the row of each dimension, the interpolated features are shown in the descending order of importance for classification. On most occasions, an attempt was made to explain the difference between the healthy and diseased image categories. Therefore, the features were interpolated from the left (diseased) to the right (healthy).

Higher values of $\beta$ appeared to remove the high-frequency features; however, it became difficult to distinguish the differences between individual features. A relatively good separation was observed at 1000. However, as can be seen from Figure 18, a $\beta$ value of 1000 resulted in a less accurate classification. <u>Thus, a very high $\beta$ value may force explainability while compromising accuracy.</u>

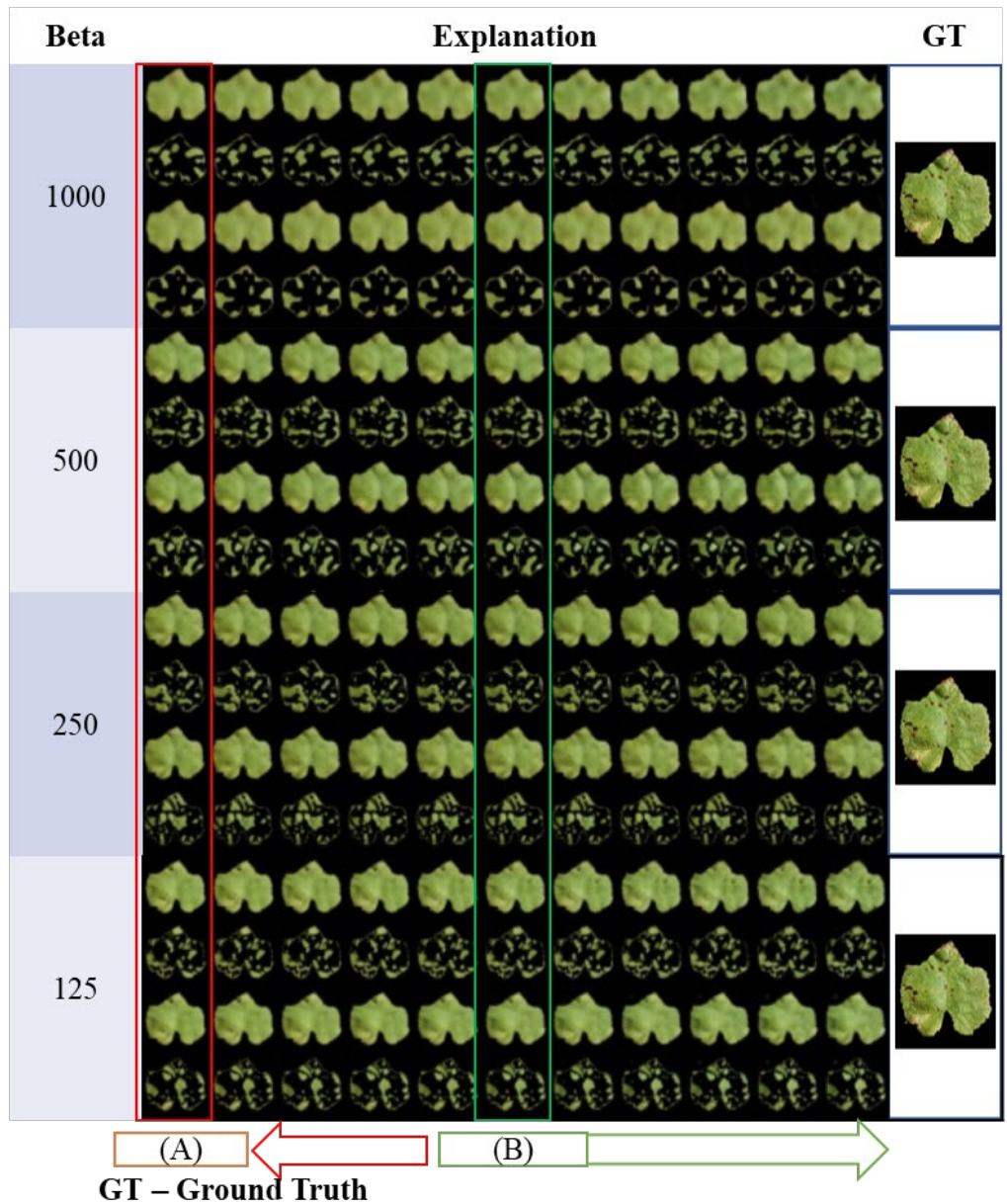

*Figure 18 Effect of $\beta$ on the explanation*

### 3.5. Effect of Dimensionality on Total Correlation, Reconstruction Loss, and Accuracy of ECLF

To test the effect of dimensionality on reconstruction, 20, 40, 80, 160, and 320 dimensions were used in the latent vector. The iteration, where the result of $L_{rc} + TC + DKL$ was the minimum, was selected because these factors are directly related to the variational autoencoder loss function; note that $\beta$ and $\gamma$ were kept constant.

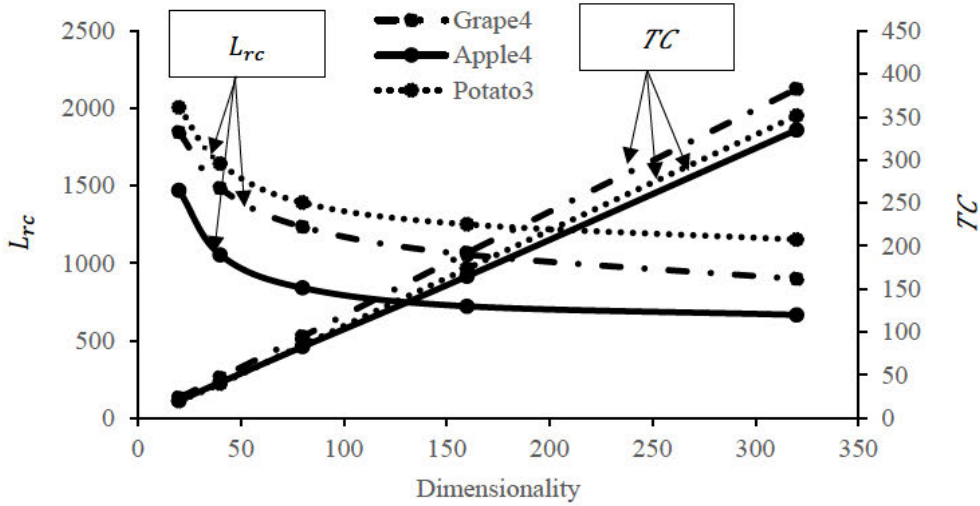

*Figure 19 Effect of dimensionality on TC and $L_{rc}$*

Figure 19 shows how the reconstruction error decreases with the dimensionality and how the total correlation increases with the dimensionality. The reconstruction error has a logarithmic relationship with the dimensionality, whereas the total correlation has a linear relationship with the dimensionality (Figure 19). This was attributed to the relaxation of the information bottleneck, which forces the variables to become independent. Theoretically, it is possible for the VAE to achieve very high reconstruction quality given equal or higher dimensions required for data representation (Dai & Wipf, 2019), however, this may adversely affect the explainability of the system since the classifiable features are more distributed in higher dimensions. Even though there is a tradeoff between $L_{rc}$ and $TC$ in lower dimensions, it appears to be more advantageous for increasing the dimensionality until $L_{rc}$ reaches flat regions. It can be seen that $TC$ increases with dimensionality (Figure 19), which means that low dimensionality acts as a major supportive factor for the reduction of $TC$ (Higgins et al., 2016).

The classification accuracy appears to increase with the dimensionality, as expected (Figure 13) for the Apple4 and Grape4 datasets, which may also be due to the relaxation of the information bottleneck, which encourages an increase in the number of classifiable features. Moreover, because a VAE performs a type of principal component analysis (Rolinek et al., 2019), this can be considered as an increase in the number of principal components in the explanations. The flattening curve of accuracy appears to support this hypothesis. The Potato3 dataset shows high accuracy, even in low dimensions, meaning that 40 dimensions appear to be sufficient. The potato dataset has prominent features such as large lesions and color changes (Figure 13) for classification, which may be the reason for the high accuracy of the dataset in low dimensions. When class differences were considered, the apple class seemed to reduce $L_{rc}$ faster than the other two classes. Considering that this class contains high-frequency information (e.g., spots and sharp lines; Figure 13), which may be deleted in lower dimensions, the dataset generally requires more dimensions than the potato dataset. As Figure 16 and 17 show, reduced dimensions can increase independence, compromising the accuracy and reconstruction quality in very low-dimensional scenarios.

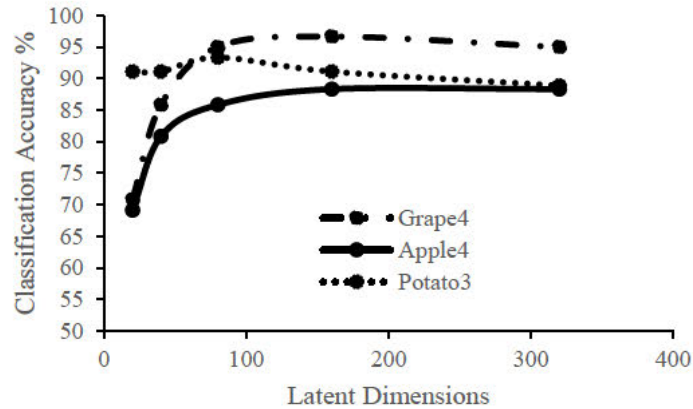

*Figure 20 Effect of latent dimensionality on accuracy*

Figure 20 shows how the classification accuracy changes with the dimensionality of the latent vector.

### 3.6. Effect of Dimensionality on the Visual Decision Explanations of ECLF

In this section, we focus on the effect of dimensionality on visual decision explanations. In Figure 21, the visualization was conducted in a manner similar to that in Figure 18, and shows that, when the dimensionality increases, the images become closer to the ground truth image. However, lower dimensions use more globally distributed features of an image, such as color. In contrast, with higher dimensions in the latent vector, the features tend to move toward local features, such as lesions and local color changes.

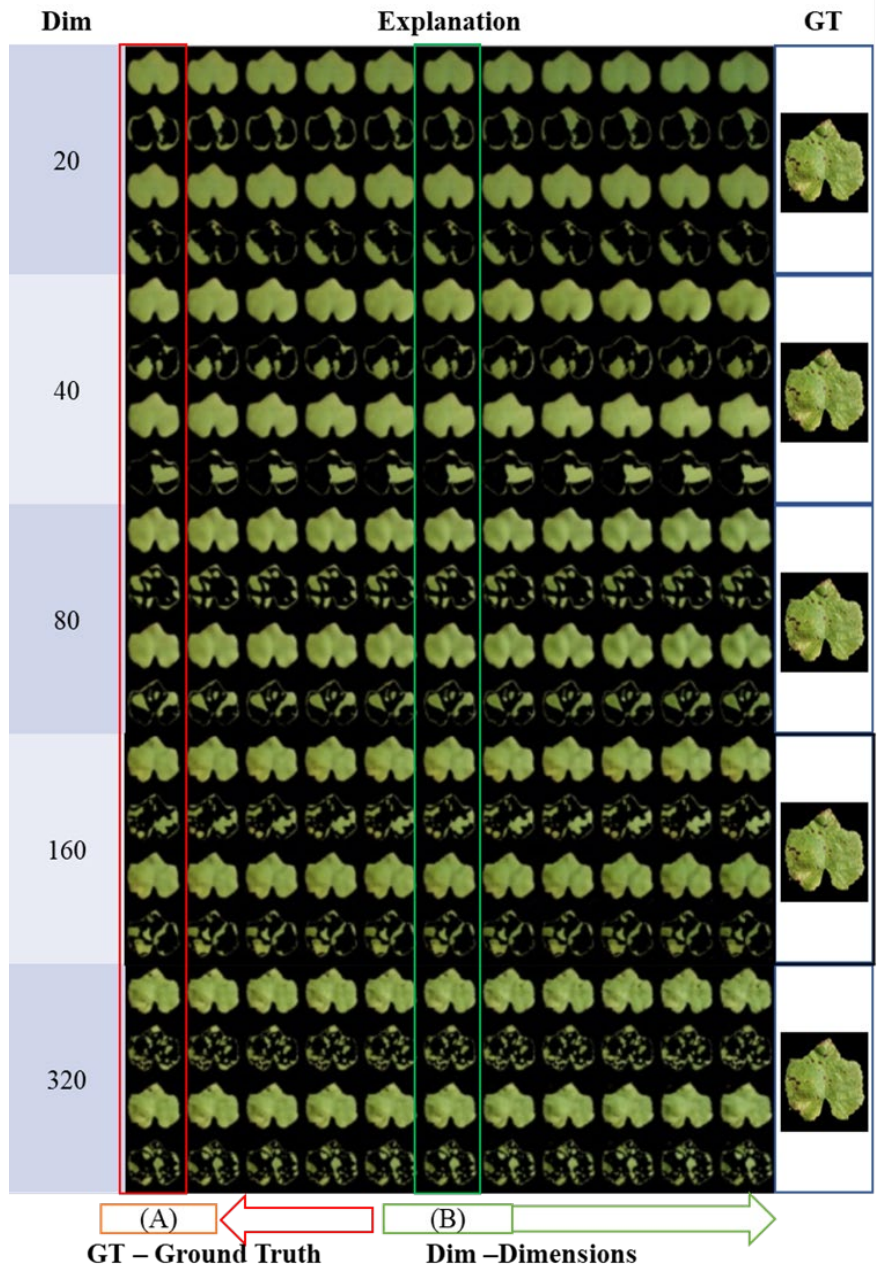

*Figure 21 Feature interpolation visualizations for grape late blight to healthy*

At 20 dimensions, the images clearly showed a color change from yellow to green, in addition to changes in shape. In this case, explainability becomes easy to achieve, but the reconstruction quality, reconstruction accuracy, and classification accuracy are compromised. In contrast, in higher dimensions, the accuracy and the amount of reconstruction increases, but explainability in the form of both disentanglement and number of dimensions is compromised.

### 3.7. Experimental Results of ECLF-CS Classification Accuracy

Using ECLF-CS, the classification accuracies for the Apple2, Grape2, and Potato2 datasets were 98.3, 98.3, and 100.0%, respectively, which were higher than those of ECLF. In particular, the accuracy was higher than that in the Apple2 dataset, which was attributed to its class-specific training, and the classifier only had to handle two classes. Even with a very high number of iterations (1,400,000), the classification accuracies were 100.0, 96.8, and 100.0% for the Apple2, Grape2, and Potato2 datasets, respectively, implying that no significant changes were observed.

### 3.8. ECLF-CS Feature Visualization

A 160-dimensional VAE, trained for 1,400,000 iterations, was used in this experiment to obtain the ECLF-CS features, using the Apple4 and Grape4 classes. As previously mentioned, the lowest loss point, where $L_{Rc} + TC + DKL$ was the minimum was used and two classes were visualized by dropping the other class in the visualization. For example, while the diseased side of the encoder produces a diseased vector, the decoder produces an image that is passed to it by the diseased vector. Therefore, there can be some information loss; however, what the decoder is producing is what information classifier used for the classification. The healthy side also tries to produce an image that is close to a healthy image. These images show us what the VAE sees in the latent space (Figure 22).

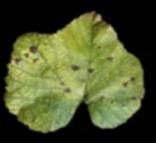

| | Lowest Loss Point | 1,400,000 iterations |
|---|---|---|
| Original | 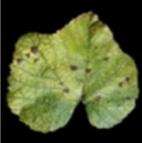 | 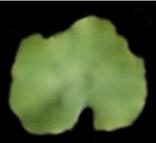 |
| Reconstructed Healthy | 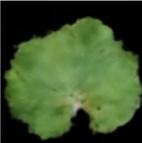 | 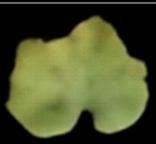 |
| Reconstructed Diseased | 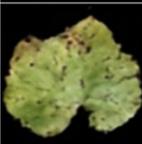 | 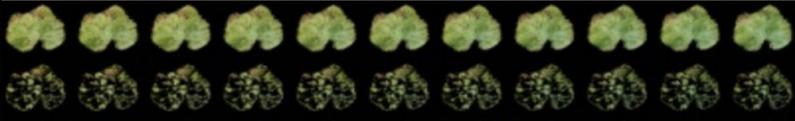 |

*Figure 22 Reconstruction of healthy and diseased images*

Figure 22 shows the difference between the reconstructed diseased and healthy images at the lowest loss point and after 1,400,000 iterations. ECLF-CS also seems to follow multiclass classification because a higher number of iterations provides better visualization.

### 3.8.1 important Feature Visualization for ECLF-CS

| Feature | Feature Interpolation Visualization | Original |
|---|---|---|
| Grape Feature 1 | 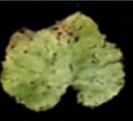 | 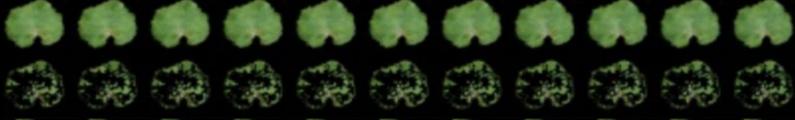 |
| Grape Feature 3 | 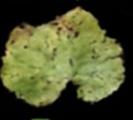 | 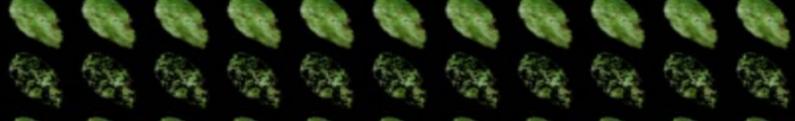 |
| Apple Feature 1 | 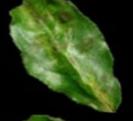 | 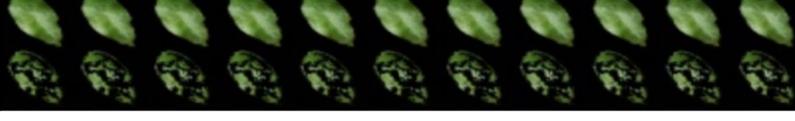 |
| Apple Feature 4 | 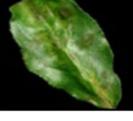 | |

*Figure 23 Important feature visualization for ECLF-CS*

In Figure 23, features are class-specific; however, in the classification stage, they are used to differentiate between classes; thus, it shows which part of the feature is considered as another class in the classification. The features of two-class classification can clearly be divided into classes A and B. Grape feature 1 belongs to the grape late-blight class, and grape feature 3 belongs to the healthy grape class. Apple features 1 and 4 of apple scab, are belong to the diseased and healthy apple classes respectively.

In ECLF, the features are trained to cross the classes during VAE training, and therefore, it is easy to understand how the features behave. However, in ECLF-CS, the features are not trained to cross the classes, and so, the feature variations must be visualized within the class images.

|  | ECLF-CS | ECLF |
|---|---|---|
| Original | 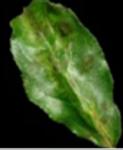 | 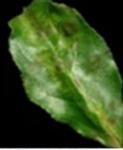 |
| Reconstructed Healthy | 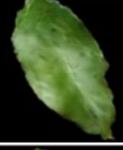 | 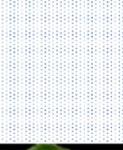 |
| Reconstructed Diseased | 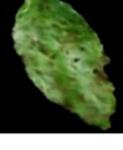 | 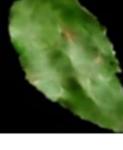 |

*Figure 24 Reconstruction comparison between ECLF and ECLF-CS*

Figure 24 shows a reconstruction comparison between the ECLF and ECLF-CS features, with the ones in ECLF-CS appearing to be of better quality, which may be due to the class-specific training that we conducted during the VAE training stage. Furthermore, <u>class-specific training may also explain why the apple class showed a larger improvement in the classification accuracy. Compared with multiclass classification, two-class classification seems to handle subtle features more easily.</u> The apple dataset has diseased features that are not very prominent to the eye, such as lesions and darkening.

### 3.9. Comparison with Other Methods

The VGG-11 network [29] was used to compare the classification accuracy of ECLF and ECLF-CS. For the training of the ECLF for two classes, the same datasets that were used for ECLF-CS were employed, and the same training and testing parameters that were used in ECLF were used. It must be noted that the VGG-11 network was not exclusively optimized for any class.

*Table 2 Comparison of the classification accuracies* [*]

| Dataset | VGG-11 | ECLF | ECLF-CS |
|---------|--------|------|---------|
| **Grape4** | 97.5% | 95.0% | - |
| **Apple4** | 96.7% | 88.3% | - |
| **Potato3** | 98.9% | 88.9% | - |
| **Grape2** | 98.3% | 98.3% | 98.3% |
| **Apple2** | 91.7% | 91.7% | 98.3% |
| **Potato2** | 100.0% | 98.3% | 100.0% |

[*]For the ECLF, the accuracies are shown at a latent vector size of 320 and for ECLF-CS, the accuracies are shown at a latent vector size of 160

For the Grape4, Grape2, Apple2, and Potato2 datasets, nearly the same classification accuracies were achieved using VGG-11 and ECLF; however, ECLF showed a decrease in accuracy on the Apple4 and Potato3 datasets. <u>In the case of the Apple4 dataset, the low</u>

accuracy was attributed to the network's dependence on high-frequency features for classification. This dependency was visually confirmed for the Apple4 dataset, as shown in Figure 13. Moreover, as discussed in Section 3.5, the classification accuracy of the Apple4 dataset depends on the value of $\beta$. As such, the number of high-frequency features reduces with the increase in $\beta$. Therefore, the classification accuracy and high-frequency features were found to have a high correlation in case of the Apple4 dataset. As such, ECLF exhibits a low capacity to capture the high-frequency classifiable features that are required in the classification of the Apple4 dataset under high explainability conditions. Moreover, as mentioned in section 3.2.1, the accuracy of the Apple4 data set increased to 94.4% at 1,500,000 iterations, but this phenomenon needs further investigation. Although a higher accuracy and explainability can be assumed if the $\beta$ value is reduced during training, an increase in the correlation between individual features may occur (Figure 16), which would hamper the explainability of the classification results.

As seen in Figure 13 with sample leaves from the Potato3 dataset, and Figure 17, which shows an almost constant relationship with increasing $\beta$, the Potato3 dataset appears to be less dependent on high-frequency features, which implies a low correlation with high-frequency features. The reduced accuracy in the case of the Potato 3 dataset may be attributed to the overfitting of the model in the 320-dimensional latent space. As can be seen from Figure 20, the classification accuracy of Potato3 classes can reach up to 93.3% in an 80-dimensional latent space. However, the ECLF showed competitive performance in the two-class classifications with the VGG-11 network. This may be because only two classes were used in the case of a dataset that allowed additional classifiable information for each class.

ECLF-CS showed competitive performances on the VGG-11 dataset and ECLF on the Grape2 and Potato2 datasets, and ECLF-CS shows improved performance on the Apple2 dataset, which was attributed to the class-specific training that was performed during the training of ECLF-CS. Although the reasons for the performance differences of ECLF and performance improvements of ECLF-CS on specific datasets compared to VGG-11 were discussed, further studies are needed to verify the stated conjectures.

## 4. Conclusions

This study details a new approach that deviates from conventional approaches for important area visualization to explain the underlying reasons that lead to a classification based on feature variations. The proposed network was trained with datasets extracted from the PlantVillage dataset and achieved acceptable accuracy with high explainability. In the future, this algorithm can be used to identify new disease symptoms that human evaluators may otherwise miss. There are some limitations of our study, including the low quality of visualization and reduced classification accuracy of the ECLF which must be addressed in future studies.

**Declaration of Competing Interest**

The authors declare that they have no known competing financial interests or personal relationships that could have influenced the work reported in this paper.

**Acknowledgments**

The authors would like to express their deep gratitude to Dr. TAKEYA Masaru for his assistance in performing this research. The authors would like to thank the Plant Protection Station of the Ministry of Agriculture, Forestry and Fisheries of Japan (MAFF), the Hokkaido Research Organization (HRO), the Tokachi Federation of Agricultural Cooperatives, the Center for Seeds and Seedlings (NCSS), the Institute of Vegetable and Floriculture Science (NIVFS), the Institute for Agro-Environmental Sciences (NIAES), and the Hokkaido Agricultural Research Center (HARC), Institute for Plant Protection (NIPP) of NARO for their invaluable cooperation. In this research work we used the NARO AI Supercomputer "Shiho".

**Funding**

Part of this research was supported by NARO, "Research project for technologies to strengthen the international competitiveness of Japan's agriculture and food industry, grant number: Mo_B_Bou_1" and a Project of the Bio-oriented Technology Research Advancement Institution (Research Program on Development of Innovative Technology, project ID: 01022C).